\documentclass[10pt, journal]{IEEEtran}
\IEEEoverridecommandlockouts

\usepackage{amssymb,amsmath,mathptmx}
\usepackage{algorithm,algorithmic}
\usepackage{epsfig,graphicx,subfigure}
\usepackage{times,array,url,color,endnotes,caption}
\pdfoutput=1
\begin{document}
\title{\LARGE \bf Real-time 3D scene description using Spheres, Cones and Cylinders }

% You will get a Paper-ID when submitting a pdf file to the conference system
\author{Kristiyan Georgiev,  Mo'taz Al-Hami,  Rolf Lakaemper%
\IEEEcompsocitemizethanks{\IEEEcompsocthanksitem All authors are with the Department of Computer and Information
Science, Temple University, Philadelphia, PA, 19107, USA.\protect\\
The project is sponsored by the grant ARRA-NIST-10D012 of the National Institute of Standards and Technology (NIST). Email: lakamper@temple.edu
}% <-this % stops a space
}

\IEEEpubid{978-1-4799-2722-7/13/\$31.00~\copyright~2013 IEEE}

\maketitle

\begin{abstract}
The paper describes a novel real-time algorithm for finding 3D geometric primitives (cylinders, cones and spheres) from 3D range data.
In its core, it performs a fast model fitting with a model update in constant time (O(1)) for each new data point added to the model.
We use a three stage approach.
The first step inspects 1.5D sub spaces, to find ellipses.
The next stage uses these ellipses as input by examining their neighborhood structure to form sets of candidates for the 3D geometric primitives.
Finally, candidate ellipses are fitted to the geometric primitives.
The complexity for point processing is O(n); additional time of lower order is needed for working on significantly smaller amount of mid-level objects. This allows the approach to process 30 frames per second on Kinect depth data, which suggests this approach as a pre-processing step for 3D real-time higher level tasks in robotics, like tracking or feature based mapping.
\end{abstract}

\begin{IEEEkeywords}
3D Features, Robot Vision, Segment, Circle, Ellipse, Sphere, Cylinder, Cone, Object Detection, Real-Time.
\end{IEEEkeywords}

%====================================================================
\section{Introduction} \label{sec:introduction}
%====================================================================d
In recent years, most range-sensing based algorithms in robotics, e.g. SLAM and object recognition, were based on a low-level representation, using raw data points. The drawbacks of such a representation (high amount of data, low geometric information) limits the scalability when processing 3D data. In 3D cases, it is more feasible to base algorithms on mid-level geometric structures. Mid-level geometric structures, such as planar patches, spheres, cylinders and cones can be used to efficiently describe a 3D scene. A recent case of how planar patches can describe a 3D scene is presented in (\cite{Georgiev:IROS:2011},\cite{poppinga2008}). Inspired by these approaches, this paper explains how to find conic features of a scene using cylinders, cones and spheres.

The goal of our work is to extract conic objects from 3D data-sets in real time. Such extraction can be used as a real-time pre-processing module for 3D feature based tasks in robotics. This includes scene analysis, SLAM, etc. based on 3D data. For example, in a recent work which deals with automatic reconstruction of buildings, \cite{MuseumCSG} expresses the need for fast geometric primitive recognition algorithm that goes beyond planes but includes cylinders and spheres, for a more accurate building representation.

We use a model fitting approach; the speed is gained by a $O(1)$ model update and dimensionality reduction of 2.5D to 1.5D (a 1D height-field, i.e. a single scan row of a range scan). In total, this achieves a complexity of $O(N)+O(M)$, with $N$ number of data points and $M$ being number of 2D ellipses, which in average is a low constant $k$ times the number of scan lines, i.e. $M = k\sqrt{N}$. This totals to a complexity of $O(N)+O(k\sqrt{N}) = O(N)$.
The speed achieved by this algorithm allows for processing 30 frames per second on 320x240 Kinect data on a 2.8GHz desktop computer, implementation in Java.

\IEEEpubidadjcol

The paper is organized as follows: after comparison to related work (Sec.\ref{sec:relatedWork}), we give an overview of the approach (Sec. \ref{sec:overview}), followed by the ellipse extraction (Sec. \ref{sec:ellipseExtraction}). Sec. \ref{sec:objectExtraction} covers the computation of the ellipse-neighborhood structure, followed by the 3D conic object detection. We prove the applicability to higher level robotic tasks and highlight some properties of the algorithm with experiments (Sec. \ref{sec:experiments}).
%-------------------------------------------------
\begin{figure}[htb]
\begin{center}
\includegraphics*[ width=1.0 \linewidth]{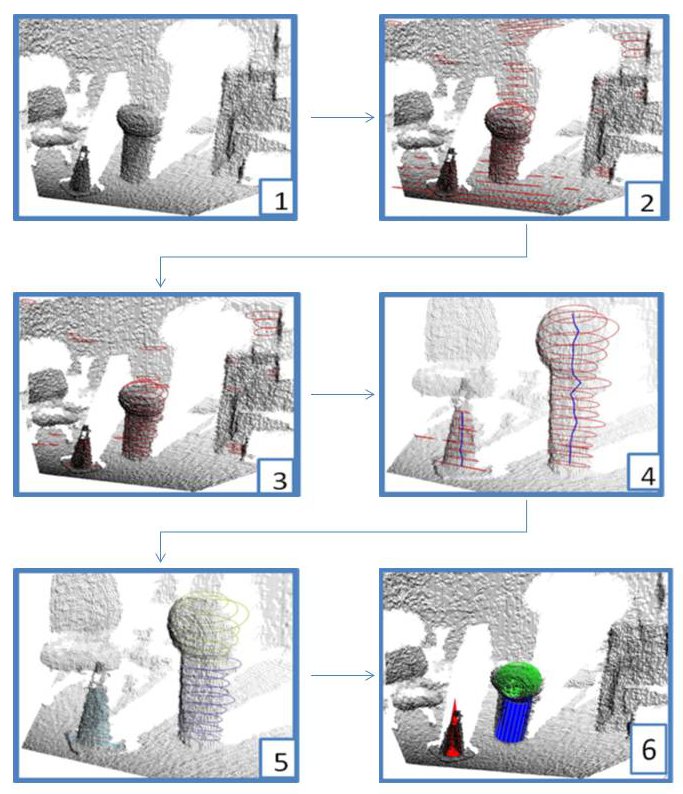}
\end{center}
\caption{\footnotesize Flow diagram showing: 1) Input data, 2) Result of ellipse fitting, 3) After removal of elongated ellipses, 4) Ellipse candidate sets based on proximity. The blue line shows connected components. 5) Resulting sets of ellipses forming different conical objects 6) Fitted 3D models.  }
\label{fig:flowChart}
\end{figure}
%-------------------------------------------------

%====================================================================
\section{Related Work}  \label{sec:relatedWork}
%====================================================================
Object detection and tracking has been long performed in robot-vision, yet most approaches process 2D RGB images. We will not compare to these approaches, but limit the related work to detection based on range data.
The problem of sphere detection has been solved in various ways and to the best of our knowledge, all previous approaches work on point clouds directly. None of these
approaches make use of a two step solution (points to ellipses, ellipses to spheres), which reduces dimensionality of the problem from 3D to  $1.5$D subsets (horizontal scan lines with distance data).

%-------------------------------------------------
\begin{figure*}[tb]
\begin{center}
\includegraphics*[ width=0.25 \linewidth]{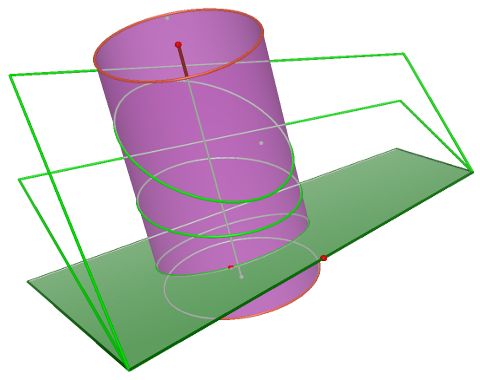}
\includegraphics*[ width=0.25 \linewidth]{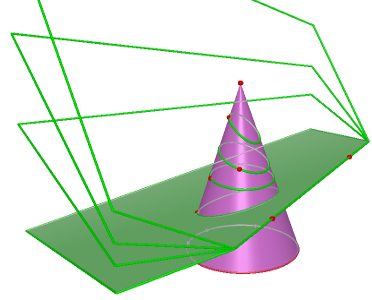}
\includegraphics*[ width=0.25 \linewidth]{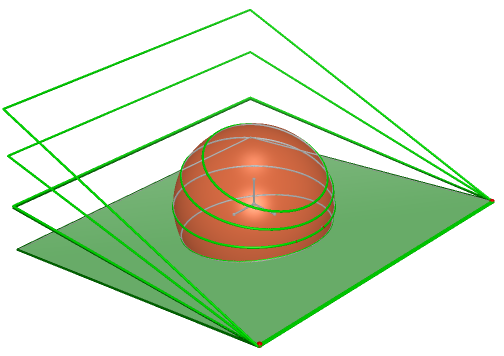}
\end{center}
\caption{\footnotesize Intersection between scanning planes (marked green) and geometric primitive (cylinder, cone and sphere) produces conic curves (ellipses, marked green) with co-linear center points. }
\label{fig:cylinder}
\end{figure*}
%-------------------------------------------------

The advantage of the two step solution is, that the higher amount of data, that
comes e.g. with 3D data, is immediately reduced: the data is decomposed into sets of lower dimensional spaces, analysis is performed in this lower dimensionality. The transition to higher dimensions is then performed on mid-level representation, here: ellipses, significantly reducing the data elements to be analyzed.

An approach that works directly on 3D point data, described in \cite{5354683}, uses surface normal vectors and forms clusters of points with similar direction.
The clustering decomposition in \cite{5354683} stays in the 3D space of the original data, which is a major difference to our approach.

In general, data modeling with pre-defined structures can also be solved by the Expectation Maximization (EM) approach. The system in \cite{lakaemper06} uses a split-and-merge extended version of EM to segment planar structures (not conic elements) from scans. It works on arbitrary point clouds, but it is not feasible for real time operation. This is due to the iterative nature of EM, including a costly plane-point correspondence check in its core. An extension to conic elements would increase the complexity even further.

A popular solution to fit models to data is Random Sample Consensus (RANSAC, \cite{RANSAC1981}). Being based on random sampling, it achieves near-optimal results in many applications, including linear fitting (which is the standard RANSAC example). However, RANSAC fails to model detailed local structures if applied to the entire data set, since, with small local data structures, nearly the entire point set appears as outliers to RANSAC. \cite{Weingarten03} applies RANSAC on regions created by a divide and conquer algorithm. The environment is split into cubes (a volume-gridding approach). If precise enough, data inside each volume is approximated by planes, and coplanar small neighboring planes are merged. This approach has similarities to ours, in the sense that it first builds small elements to create larger regions. However, their split stays in the third dimension, while our split reduces the dimensionality from 3D to 1.5D, which makes the small-element generation, i.e. ellipse, a faster operation. We have a more detailed comparison to RANSAC in section \ref{sec:RANSAC}.

Ellipse extraction is a crucial step in our solution. There are many different methods for detection and fitting ellipses. \cite{Halir98numericallystable} has done an extensive overview of different methods for detection and fitting ellipses. These methods range from Hough transform \cite{Yip19921007, Wu19931499}, RANSAC \cite{368167}, Kalman filtering \cite{476507}, fuzzy clustering \cite{Gath1995727}, to least squares approach \cite{FredL197956}. In principle they can be divided into two main groups: voting/clustering and optimization methods. The methods belonging to the first group (Hough transform, RANSAC, and fuzzy clustering) are robust against outliers yet they are relatively slow, require large memory but have a low accuracy.
The second group of fitting methods are based on optimization of an objective function, which characterizes a goodness of a particular ellipse with respect to the given set of data points. The main advantages of these methods are their speed and accuracy. On the other hand, the methods can fit only one primitive at a time (i.e. the data should be pre-segmented before the fitting). In addition, the sensitivity to outliers is higher than in the clustering methods. An analysis of the optimization approaches was done in \cite{765658}. There were many attempts to make the fitting process computationally effective and accurate. Fitzgibbon et al. proposed a direct least squares based ellipse-specific method in \cite{765658}.

Our approach to ellipse extraction belongs to the second group as a type of optimization, yet with the advantage of finding multiple objects at one time. It was motivated by the work of \cite{Georgiev:IROS:2011,poppinga2008}. In there, a region growing algorithm is proposed on point clouds, testing an optimal fit of updated planes to the current region. This paper extends the work of \cite{Georgiev:IROS:2011} from planes to spheres, cylinders and cones and overcomes the drawbacks, because of the use of a numerically stable ellipse fitting and the fast $O(1)$ model update.

%====================================================================
\section{Method Overview}  \label{sec:overview}
%====================================================================
The guiding principal of our work is to look for mid-level geometric elements (MLEs) in lower dimensional subspaces and then to extend the data analysis to the missing dimensions using these MLEs. The advantage is, that the number of MLEs is significantly smaller than the number of data points. In addition modeling of mid level elements in lower dimensions is an easier task.
In our specific case of finding spheres, cylinders and cones, we are traversing $1.5$-dimensional subsets (horizontal scan lines with distance data) of $3D$ data. The modeling of geometric primitives reduces to finding ellipses. The intersection of such primitives with the scanning plane consists of ellipses, see Fig. \ref{fig:cylinder}. Therefore, for each scan line, we traverse its data points iteratively, and try to fit ellipses. We determine maximal connected subsets of each scan line, which fit ellipses within a certain radius interval, and under an error threshold $T_{\epsilon}$. This process is performed in an iterative way, updating currently found ellipses in $O(1)$ by adding the next candidate point.  The ellipses $E_i^s=(c,r_1,r_2)$ in scan line $s$ are stored as center point $c$ and radius $r_1$ and radius $r_2$. Traversing each scan line therefore leaves us with a set of ellipses $\mathcal{E}=\{E_i^s\}$. The set of all ellipses $\mathcal{E}$ is decomposed into subsets of connected components based on ellipse center proximity.

After ellipse detection, it holds for cylinders and cones, that the center points of participating ellipses' center points are collinear, see Fig. \ref{fig:cylinder}, and concyclic for spheres. It is therefore sufficient, to scan the connected components for being collinear or concyclic in order to find model candidates. Please note, that this approach reduces the effort, again, to a simple line or ellipse fitting, yet now even in the significantly smaller space of center points. Such lines and ellipses from the center point space are again found with an iterative $O(1)$ update approach.

Cones, spheres and cylinders are characterized by the change of radius along the vertical axis. Hence we can determine, if a connected component of ellipses constitutes one of the models.

In conclusion, we find simple geometric models, ellipses, in a subspace of reduced dimensionality (1.5D), then perform low dimensional fitting again in a 1.5D space determined by the model. This split leads to an addition instead of multiplication of run times of the sub-tasks, which makes the approach fast.  Please see Fig. \ref{fig:flowChart} for an illustration of the approach.

%====================================================================
\section{Ellipse Extraction}  \label{sec:ellipseExtraction}
%====================================================================
A well-known non-iterative approach for fitting ellipses on segmented data (all points belong to one ellipse) is Fitzgibbon's approach described in \cite{765658}. It is an optimization approach, which uses a (6x6) scatter matrix $S$ as shown in Eq. \ref{eq:S}, to solve the problem of finding the six parameters which describe an ellipse. A more detailed explanation can be found in \cite{765658} and is out of the scope of this paper. However, it is important that $S$ has entries $S_{x^p y^q} = \sum_{i=1}^N{x_i^p y_i^q}$, where $p$ and $q$ are integers. $S$ has to be decomposed into eigenvalues and eigenvectors. The 6D vector-solution to describe an ellipse is the eigenvector corresponding to the smallest eigenvalue. Though the algorithm is robust and effective, it has two major flaws. First, the computation of the eigenvalues of Eq. \ref{eq:S} is numerically unstable. Second, the localization of the optimal solution is not correct all the time as described by \cite{Halir98numericallystable}.

\begin{algorithm}
\begin{algorithmic} [frame=single]  \algsetup{indent=2em}
\WHILE{$(nextPoint = Points.next()) \neq NULL$}
\STATE $ addPoint(nextPoint) $
\STATE $ fitModel() $
\STATE $ calculateError() $
\IF {$error > THRESHOLD$}
	\STATE $ removeLastPoint() $
	\STATE $ saveCurrentModel() $
	\STATE $ startNewModel() $
\ENDIF
\ENDWHILE
\label{alg:ellipseFitAlgo}
\end{algorithmic} \caption{extractEllipses(Points, THRESHOLD)}
\end{algorithm}

To overcome the drawbacks \cite{Halir98numericallystable} splits $S$ into three matrices ($S_1, S_2, S_3$), which are (3x3), see Eq. \ref{eq:Scomosite}a,b. With this decomposition, \cite{Halir98numericallystable} achieves the same theoretical result, but with improved numerical stability and without the localization problem.

However, both algorithms, \cite{Halir98numericallystable} and  \cite{765658} only find a single ellipse fit on a given set of pre-segmented points. We propose a solution based on Fitzgibbon's approach and the numerical improvement made by \cite{Halir98numericallystable} to be able to handle non segmented data, i.e. data containing points not necessarily belonging to an ellipse and possibly containing disjoint subsets of points belonging to different ellipses. Our approach finds multiple ellipses in a single pass. We achieve this by performing an ellipse model update in $O(1)$ and making use of the natural point order defined by the range sensor, see Algorithm 1.
To perform the $O(1)$ update we store each component of the sum ($S_{x^p y^q}$) required to form matrices $S_1, S_2, S_3$. This means when adding a new point to the existing model, we only have to perform addition to each of the sum terms, which is a constant time operation. For a set of $N$ points the model update will be executed exactly $N$ times, our algorithm can find multiple ellipses in non segmented data in $O(N)$.

\begin{equation}\label{eq:S}
S = D^TD =
\begin{bmatrix}
S_{x^4} & S_{x^3y} & S_{x^2y^2} & S_{x^3} & S_{x^2y} & S_{x^2} \\
S_{x^3y} & S_{x^2y^2} & S_{xy^2} & S_{x^2y} & S_{xy^2} & S_{xy} \\
S_{x^2y^2} & S_{xy^3} & S_{y^4} & S_{xy^2} & S_{y^3} & S_{y^2} \\
S_{x^3y} & S_{x^2y} & S_{xy^2} & S_{x^2} & S_{xy} & S_{x} \\
S_{x^2y} & S_{xy^2} & S_{y^3} & S_{xy} & S_{y^2} & S_{y} \\
S_{x^2} & S_{xy} & S_{y^2} & S_{x} & S_{y} & S_{1} \\
\end{bmatrix}
\end{equation}

%-------------------------------------------------
\begin{figure}[htb]
\begin{center}
\includegraphics*[ width=0.39 \linewidth]{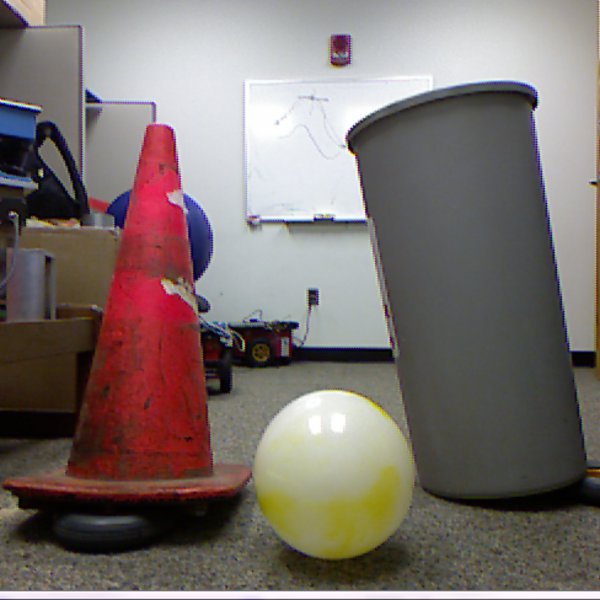}
\includegraphics*[ width=0.39 \linewidth]{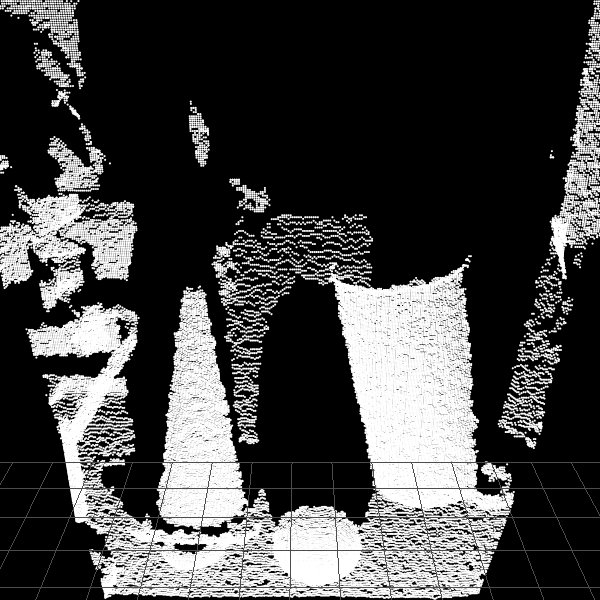}
\includegraphics*[ width=0.39 \linewidth]{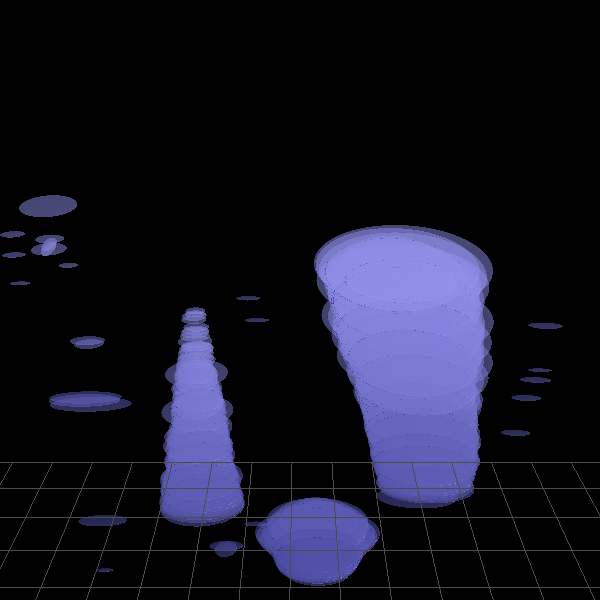}
\includegraphics*[ width=0.39 \linewidth]{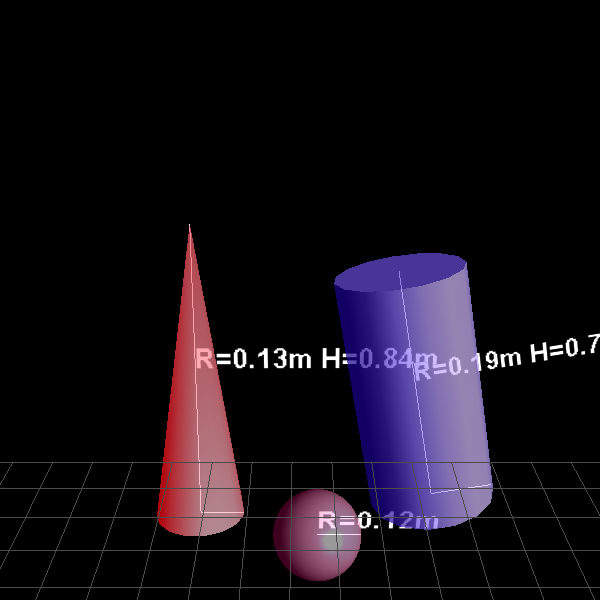}
\includegraphics*[ width=0.79 \linewidth]{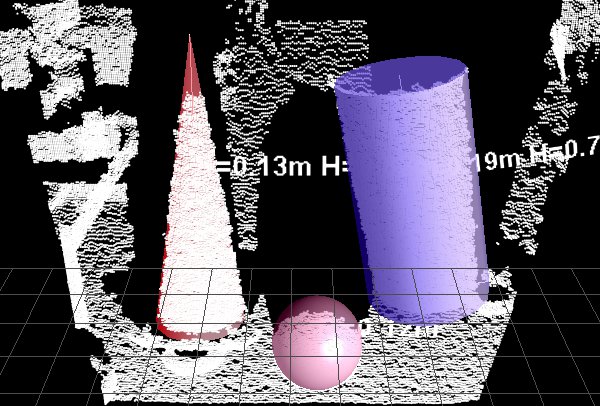}
\end{center}
\caption{\footnotesize  Object detection. Top to bottom: RGB image(not used); 3D point cloud from MS Kinect (307200 points); extracted ellipses after the pre-filter; to the right are the fitted conic objects; at the bottom are fitted conic objects and 3D point cloud. The cylinder in the picture is a cylindrical trash can. }\label{fig:3DTest_1}
\end{figure}
%-------------------------------------------------

\begin{subequations} \label{eq:Scomosite}
\begin{align}
       S &= \begin{bmatrix}  S_1 \vline S_2 \\ \hline S_2^T \vline S_3 \end{bmatrix}  &
       S_1 &= \begin{bmatrix} S_{x^4} & S_{x^3y} & S_{x^2y^2} \\ S_{x^3y} & S_{x^2y^2} & S_{xy^3} \\ S_{x^2y^2} & S_{xy^3} & S_{y^4} \end{bmatrix} \\
       S_2 &= \begin{bmatrix} S_{x^3} & S_{x^2y} & S_{xy^2} \\ S_{x^2y} & S_{xy^2} & S_{y^3} \\ S_{x^2} & S_{xy} & S_{y^2} \end{bmatrix} &
       S_3 &= \begin{bmatrix} S_{x^2} & S_{xy} & S_{x} \\ S_{xy} & S_{y^2} & S_{y} \\ S_{x} & S_{y} & S_{1} \end{bmatrix}
\end{align}
\end{subequations}

\subsection{System Limitations}
\begin{figure}[htb]
\begin{center}
\includegraphics[ width=0.3\linewidth]{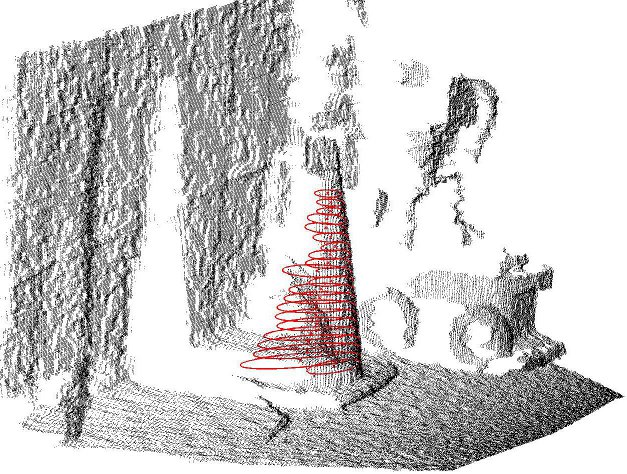}
\includegraphics[ width=0.3\linewidth]{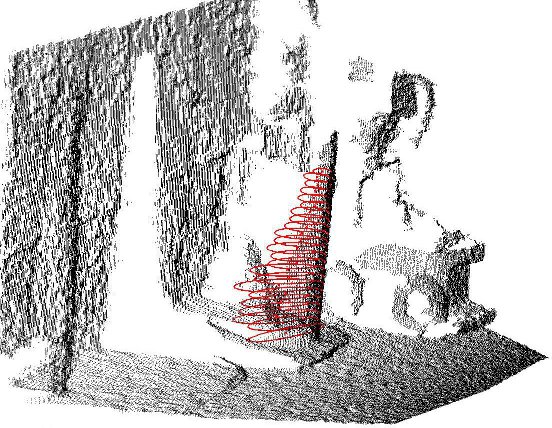}
\includegraphics[ width=0.3\linewidth]{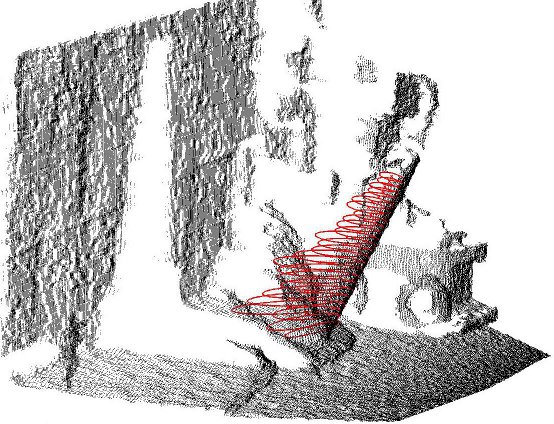}
\end{center}
\caption{\footnotesize A tilted (rotation around x-axis, tilted towards camera) parking cone, and ellipses detected (red). Top Left: tilt=5 degrees, Top Right: tilt=20 degrees, Bottom: tilt=40 degrees. In all cases, the algorithm is able to fit a cone model (not explicitly shown in figure).
}\label{fig:cone_test}
\end{figure}

Being based on finding ellipses from the conical intersection of a scanning plane with cylinders, cones and spheres it is limited with respect to tilting angles that lead to non-elliptical intersection patterns. The appearance of such patterns depend on the size of the object (cylinder) and the size and internal angle (cone). In practice, our system performs robustly on tilting angles of $>45$ degrees, see Fig. \ref{fig:3DTest_1}, yet is naturally limited when the angle becomes significantly larger. One remedy of this problem (for tilts resulting from rotation around the $z$ axis, is to also use ellipses found by intersection of vertical instead of horizontal scanning planes, i.e a $90$ degree rotated sight of the scene, which turns tilts of $>45$ into tilts of $<45$. An example of tilted objects is illustrated in Fig. \ref{fig:cone_test} and Fig. \ref{fig:3DTest_1}.

%====================================================================
\section{3D Object Extraction}  \label{sec:objectExtraction}
%====================================================================
Once all ellipses are extracted, further processing is performed on these ellipses only; the original point data is not used any longer. This significantly reduces the amount of data used (usually about $k \sqrt{n}$ times, where $\sqrt{n}$ results from the number of scan lines vs. the number $n$ of data points in the entire image, assuming for simplicity a square sized image). The ellipses are used to form 3D objects, such that centers of supporting ellipses are co-linear and the centers lie inside of an error margin $\phi$, the maximum allowed distance between neighboring ellipses. Only the first $k$ neighbors are compared to eliminate outlier ellipses due to noise. This allows for selective skipping vertically and choosing better supporting ellipses for the object.

More formally:
The 1.5D ellipse extraction creates a set $\mathcal{E}=\{E_i\}$, $E_i=(x_i,y_y,z_y,r1_i,r2_i)$ of ellipses, with center point $c_i=(x_i,y_i,z_i)$.
For our target objects, conic 3D objects, we are looking for sequences $S_j = (c_{i1},c_{i2},..,c_{ik})$ of collinear or concyclic centerpoints $c_i$.
The order of extraction imposes an order on ellipses $E_i \in S_j \subset \mathcal{E}$, they are ordered in the $z$-dimension:
\begin{equation}
i1 < i2 \Rightarrow z_{i1} < z_{i2}
 \end{equation}

%-------------------------------------------------
%-------------------------------------------------

Due to this order, we can extract collinear/concyclic center-points using the $O(|\mathcal{E}|)$ line/ellipse fitting algorithm, which in its core again has an $O(1)$ update. For details on incremental $O(n)$ line detection, please see \cite{Georgiev:IROS:2011}. Similarly, ellipse models fitting the center points are extracted, using the method described in Sec. \ref{sec:ellipseExtraction}.

\begin{figure}[htb]
\begin{center}
\includegraphics*[ width=0.5 \linewidth]{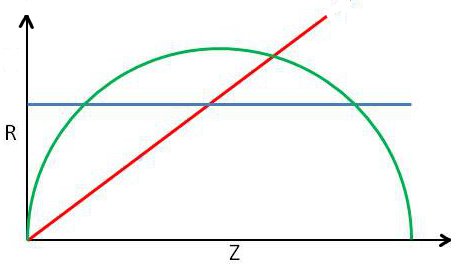}
\end{center}
\caption{\footnotesize The three curves represent ellipse radii (R-axis) along the center lines (Z-axis)  of  a cylinder (blue), cone (red) and sphere (green half circle) in (z,r) space. }
\label{fig:ZR}
\end{figure}

Once sequences of collinear center-points $S_j$ are extracted, the corresponding radii $r1_i,r2_2$ can be analyzed to detect conic objects. It is sufficient to examine the radius orthogonal to scan-lines, $r2$. Hence, object detection is performed in the $(z,r2)$, or short: $(z,r)$, space, see Fig. \ref{fig:ZR}.

%-------------------------------------------------
%-------------------------------------------------

\subsection{Cylinder}
For a cylinder, $r$ is constant:
\begin{equation}
f_{cylinder}(z) = r
\end{equation}
In Fig. \ref{fig:ZR}, this is the straight line parallel to the Z-axis. In practical application, we allow for 5 degrees deviation from parallel due to noise.
Please see also Fig. \ref{fig:cylinder}, Left.

\subsection{Cone}
For a cone, there is a linear dependency between r and z:
\begin{equation}
f_{cone}(z)=\alpha z
\end{equation}
In the practical application, the slope of the line should be greater than a small angular threshold $\theta$  (for Kinect data $\theta=5 deg$).
See Fig. \ref{fig:ZR} and Fig. \ref{fig:cylinder}, Center.
\subsection{Sphere}
For a sphere, the radii describe a circle:
\begin{equation}
f_{circle}(z)=\sqrt{r^2-(z-z_i)^2}
\end{equation}

\begin{figure}[htb]
\begin{center}
\includegraphics[ width=0.49\linewidth]{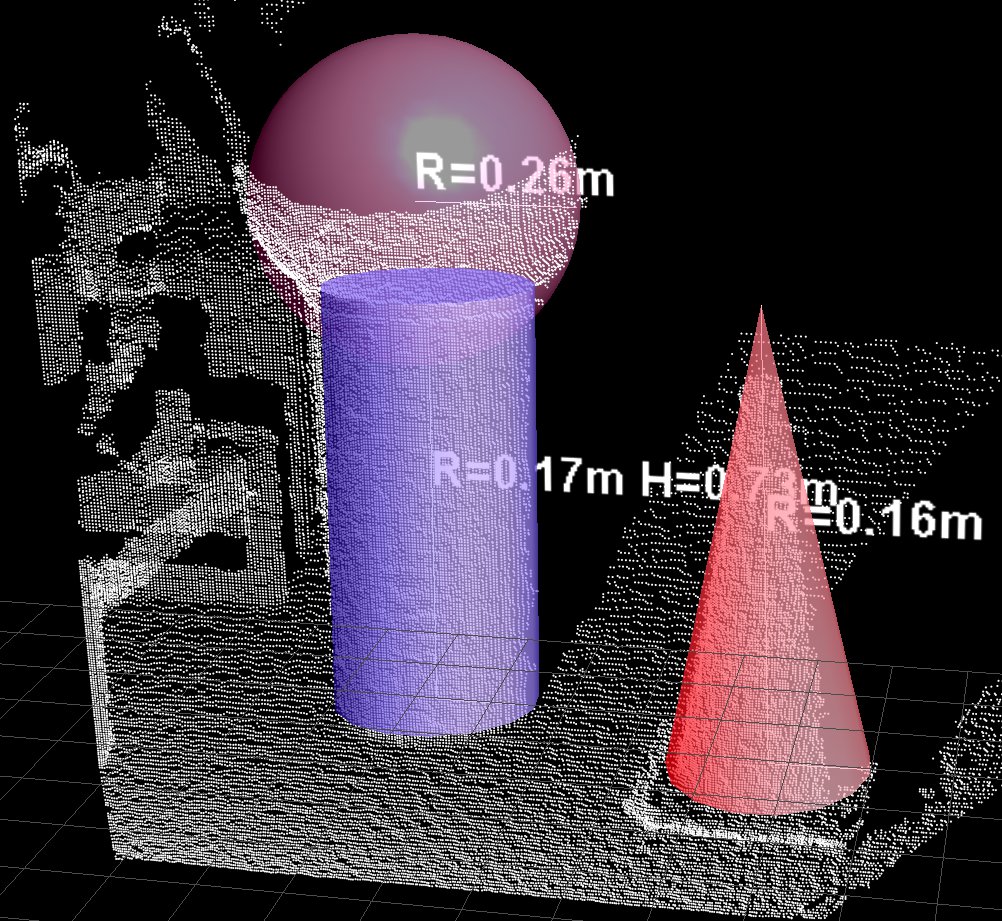}
\end{center}
\caption{\footnotesize Results of the algorithm showing fitted 3D geometric models. The sphere placed on top of a cylinder is successfully recognized as a separate object. }\label{fig:3DTest_2}
\end{figure}

The detection of objects and collinear sequences is performed in parallel, again using incremental line (and ellipse) fitting procedures, yet in different spaces: while the center-collinearity is detected by a line fitting algorithm in the $(x,y,z)$ space, the detection of 3D objects is performed by a line fitting (for cylinders and cones) and ellipse fitting (for spheres) in the $(z, r)$ space. The incremental order in $z$ reduces the collinearity finding to a 1.5D problem.
Please note that the same concept of incremental line/ellipse fitting is utilized to solve tasks not only in different phases of the algorithm (first, ellipse finding in scan rows, then line/ellipse finding on ellipse center points), but also in different spaces, the decomposed 1.5D data space, and the (z,r) space. Using the   $O(1)$ update principle in all cases is the main reason for the  fast performance of the algorithm. In addition, the nature of the fitting procedures automatically separates objects in composed scenes, see e.g. Fig. \ref{fig:3DTest_2}, cylinder and sphere.

%-------------------------------------------------
%-------------------------------------------------
\section{Improving the System using Kalman Filtering}

The data from the Microsoft Kinect sensor suffers from limited accuracy. The work in \cite{kinectError2012} shows that there is a direct relation between the sensor error and the measured distance ranging from few millimeters (at low range about 1m) up to 4 cm (at far range, about 5m). To improve the data accuracy, we use a Kalman filter as a supporting tool to reducing the amount of  errors in the measurement  observations.  Kalman filtering is a well known approach to reduce noise from a series of measurements observed over time for a dynamic linear system. In our experiments, we utilize the Kalman filter to increase precision, and to be able to do object tracking. Please note that a Kalman filter takes a series of measurements. Our system, due to its real-time behavior, is able to feed data into the Kalman filter at a high frequency (about 30Hz), which makes the combination of Kalman filtering with our approach feasible and useful.

%====================================================================
\section{Experiments}  \label{sec:experiments}
%====================================================================
\subsection{Basic Test: Object detection, detection speed}
We acquire 3D range data from a Microsoft Kinect sensor to demonstrate the algorithms basic functionality, object recognition, using the methods described in Sec. \ref{sec:objectExtraction}. The scene consist of a trash can (cylinder), parking cone and exercise ball (sphere) on the floor next to each other, see Fig. \ref{fig:3DTest_1}. The algorithm successfully finds the correct conical objects in the scene with a rate of 30fps for $320\times240$ Kinect depth data, implementation in JAVA on Dell 2.8GHz desktop with 8GB Ram.

%-------------------------------------------------
\begin{figure}[tb]
\begin{center}
\includegraphics[ width=0.65 \linewidth]{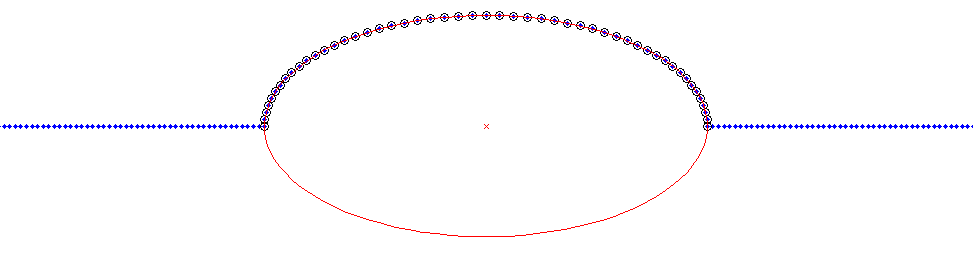}
\includegraphics[ width=0.65 \linewidth]{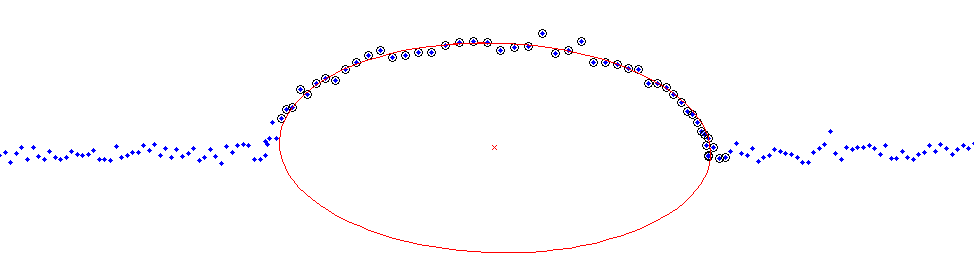}
\includegraphics[ width=0.65 \linewidth]{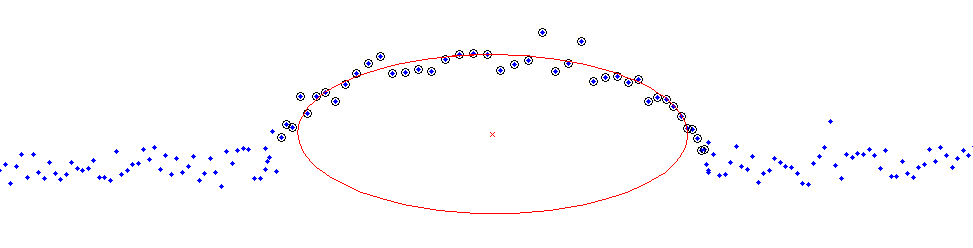}
\end{center}
\caption{\footnotesize Result of ellipse extraction in noisy, simulated data. Blue dots are input data, the fitted ellipse and its center (marked red) and supporting points (marked black). Starting from top to bottom, random noise is added with $\sigma$: 0.00, 1.0, 2.0. }
\label{fig:noiseExp}
\end{figure}
%-------------------------------------------------

\subsection{Robustness of Mid-Level Geometric Features to Noise}  \label{sec:noiseExperiments}
\begin{figure}[htb]
\begin{center}
\includegraphics*[ width=0.99 \linewidth]{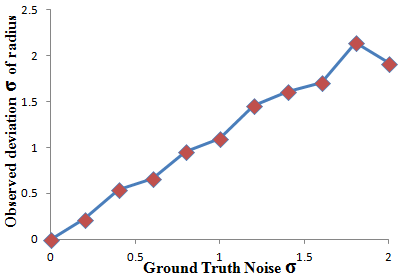}
\end{center}
\caption{\footnotesize Result of ellipse extraction in noisy, simulated data. Amount of added random noise to the data (x-axis) vs standard deviation of fitted ellipse model over 50 trials. }\label{fig:test_noise}
\end{figure}

 Robustness to noise is tested towards noise by generating a data set (see fig. \ref{fig:noiseExp}), and changing the amount $\sigma$ of Gaussian noise. For each $\sigma$, a series of 50 data sets is generated and analyzed by our algorithm to detect the ellipse. The results, see Fig. \ref{fig:test_noise}, show a benign behavior towards noise, even with a very high noise level. The variance in radius matches the variance in noise as expected (the figure shows the results without Kalman filtering. With Kalman filtering, the ellipses converge to the ground truth ellipse, as expected). Note that even with noise of $\sigma = 2.0$ the ellipse ($R_{major}=40, R_{minor}=20$) is still detected. The error threshold for each noise experiments is set proportional to the amount of noise. In real world data, this parameter can be determined by the sensor's technical specifications, and is therewith fixed.

%-------------------------------------------------
%-------------------------------------------------

%%%%%%%%%%%%%%%%%%%%%%%%%%%%%%%%%%%%%%%%%%%%%%%%%%%%%%%%%%%%%%%%%%%%%%%%%%%%%%%%%%%%%%%%%%%%%%%%%%%%%%%%%%%%%%%%%%%%%%%%%%%%%%%%%

\begin{table*}[htbp]
\centering
%\caption{Results from the used algorithm observing a cylinder at different distances ($D_{G}$). Average distance is computed from the observed measurements only ($ D_{M} $), and from observed measurements with kalman filter ($ D_{K\sigma} $).}
\subtable[No Kalman ]{
 \begin{tabular}{ | l | l | l | l | l | l | }
    \hline
$D_{G}$  & $ D_{M} $ & $ D_\sigma $ & $R_{M}$ & $R_\sigma $  & $N$ \\ \hline
1.0 & 	1.03 & 	1.08e-3 & 	0.20 & 	1.08e-3	 & 56 \\ \hline
1.5 & 	1.52 & 	1.15e-3 & 	0.17 & 	7.76e-4	 & 54 \\ \hline
2.0 & 	2.02 & 	2.11e-3 & 	0.16 & 	1.25e-3	 & 57 \\ \hline
2.5 & 	2.47 & 	2.62e-3 & 	0.16 & 	1.29e-3	 & 63 \\ \hline
3.0 & 	3.05 & 	5.16e-3 & 	0.15 & 	3.00e-3	 & 54 \\ \hline

    \hline
    \end{tabular}
\label{tab:noKalman}
}
\subtable[With Kalman]{
\begin{tabular}{ | l | l | l | l | l | l |}
    \hline
$D_{G}$  &  $D_K$ & $ D_{K\sigma} $ & $R_K$  & $R_{K\sigma}$  & $N$ \\ \hline
1.0 & 	1.03 & 	1.29e-4 & 	0.20 & 	1.47e-7 & 	56 \\ \hline
1.5 & 	1.52 & 	5.94e-5 & 	0.17 & 	1.27e-7 & 	54 \\ \hline
2.0 & 	2.02 & 	8.08e-5 & 	0.16 & 	7.43e-7 & 	57 \\ \hline
2.5 & 	2.47 & 	4.46e-4 & 	0.16 & 	6.84e-7 & 	63 \\ \hline
3.0 & 	3.05 & 	1.31e-3 & 	0.16 & 	1.91e-6 & 	54 \\ \hline

    \hline
    \end{tabular}
\label{tab:Kalman}
}
    \caption{Accuracy test. Left table: without Kalman filter. $D_G$: ground truth distance, $D_M$: mean measured distance, $D_\sigma$: standard deviation of distance measurements, $R_M$: mean measured radius, $R_\sigma$: standard deviation of radius measurements, $N$: number of measurements. Right table: with Kalman filter, labels accordingly.} \label{tab:accuracy}
\end{table*}

%%%%%%%%%%%%%%%%%%%%%%%%%%%%%%%%%%%%%%%%%%%%%%%%%%%%%%%%%%%%%%%%%%%%%%%%%%%%%%%%%%%%%%%%%%%%%%%%%%%%%%%%%%%%%%%%%%%%%%%%%%%%%%%%%%%%%%%%%
\begin{table} \footnotesize
\centering
    \begin{tabular}{ | l | l | l | l | l | l | l |}
    \hline
$D_{G}$  & $O_\%$ & $R$ & $R_\sigma$ & $R_K$  & $R_{K\sigma}$ & $N$ \\ \hline
 1.0 & 	10\% & 		 0.20 & 	9.89e-4		 & 0.20 & 	1.78e-4 & 	61 \\ \hline
 1.0 & 	20\% & 		 0.17 & 	7.39e-4		 & 0.17 & 	2.17e-4 & 	57 \\ \hline
 1.0 & 	30\% & 		 0.14 & 	6.58e-4		 & 0.14 & 	1.11e-4 & 	61 \\ \hline
 1.0 & 	40\% & 		 0.11 & 	3.16e-3		 & 0.11 & 	3.71e-4 & 	61 \\ \hline
 1.5 & 	10\% & 		 0.16 & 	5.72e-4		 & 0.16 & 	1.10e-4 & 	58 \\ \hline
 1.5 & 	20\% & 		 0.16 & 	4.03e-2		 & 0.16 & 	4.59e-3 & 	67 \\ \hline
 1.5 & 	30\% & 		 0.16 & 	3.26e-2		 & 0.17 & 	4.21e-3 & 	63 \\ \hline
 1.5 & 	40\% & 		 0.13 & 	1.27e-2		 & 0.13 & 	1.88e-3 & 	62 \\ \hline
 2.0 & 	10\% & 		 0.16 & 	9.03e-4		 & 0.16 & 	4.74e-4 & 	57 \\ \hline
 2.0 & 	20\% & 		 0.16 & 	7.08e-4		 & 0.16 & 	2.60e-4 & 	59 \\ \hline
 2.0 & 	30\% & 		 0.13 & 	1.21e-2		 & 0.14 & 	2.88e-4 & 	68 \\ \hline
 2.0 & 	40\% & 		 0.13 & 	1.77e-2		 & 0.13 & 	3.02e-3 & 	52 \\ \hline

    \hline
    \end{tabular}
    \caption{Occlusion experiment of a cylinder($R=0.2m$) at different distances ($D_{G}$), with and without Kalman filtering. $O$: occlusion percentage (10-40), the algorithm does not handle occlusions $\geq$ 50\%. $R$: measured radius, $R_\sigma$: standard deviation radius, $R_K$, $R_{K\sigma}$: with Kalman filter, $N$: number of measurements.}
    \label{tab:Occlusion}
\end{table}

%%%%%%%%%%%%%%%%%%%%%%%%%%%%%%%%%%%%%%%%%%%%%%%%%%%%%%%%%%

\begin{table} \footnotesize
\centering
    \begin{tabular}{ | l | l | l | l | l | l |}
    \hline
$V_{G}$  & $V{M}$ & $V_\sigma$ & $V_K$ & $V_{K\sigma}$  & $N$ \\ \hline
0.21 & 	0.22 & 	7.79e-2 & 	0.21& 	4.17e-2 & 97  \\ \hline
0.65 & 	0.64 & 	3.18e-1 & 	0.61& 	1.81e-1 & 63  \\ \hline
0.71 & 	0.72 & 	2.19e-1 & 	0.72& 	1.23e-1 & 27  \\ \hline
1.09 & 	0.84 & 	4.46e-1 & 	0.93& 	4.08e-1 & 22  \\ \hline
1.54 & 	1.25 & 	5.71e-1 & 	1.45& 	4.69e-1 & 13  \\ \hline
1.64 & 	1.61 & 	9.41e-1 & 	1.36& 	9.25e-1 & 6   \\ \hline
2.66 & 	2.51 & 	8.79e-1 & 	2.38& 	8.35e-1 & 6   \\ \hline

    \hline
    \end{tabular}
    \caption{Observations of a moving cylinder. Estimated velocity $V_{M}$ is computed using the observed change in distance over time and is compared to the ground truth velocity ($V_{G}$), $V_\sigma$ denotes standard deviation in velocity. Similar computation is shown using the Kalman filter ($V_K$, $V_K\sigma$). $N$: number of measurements.}
    \label{tab:speed}
\end{table}

%%%%%%%%%%%%%%%%%%%%%%%%%%%%%%%%%%%%%%%%%%%%%%%%%%%%%%%%%%

\subsection{Accuracy}  \label{sec:AccuracyExperient}
The accuracy of the used approach is measured by placing a cylinder (trash can) in front of the Kinect sensor at different distances. Several observations ($N$) were made to compute the mean and standard deviation ($ D_\sigma $) for the cylinder distance to the sensor, see Table \ref{tab:accuracy}. The results in Table
\ref{tab:accuracy}\ref{tab:noKalman}
show a relatively close measured ($ D_{M} $) to the ground truth distance with a small amount of uncertainty, and at the same time as the distance from the cylinder to kinect sensor increase, the error in observing the radius ($R_{M}$) increase. Table
\ref{tab:accuracy}\ref{tab:Kalman}
shows the effect of Kalman filtering on the observations, the distance with Kalman filtering ($D_K$) is close to the one obtained by observations only but with lower variation ($R_\sigma$). The same situation occurs for the radius. The Kalman filter in this experiment reduces the standard deviation, making the observations more certain.

\subsection{Occlusion}  \label{sec:RobustnessExperient}
A cylinder of radius $r=20$cm was obstructed by different amounts at various distances to measure the robustness of this approach w.r.t. distance and occlusion, see Table \ref{tab:Occlusion}. As the occlusion amount ($O_\%$) increases, the observed radius $R$ decreases for the observations with and without Kalman. This is due to the limited and noisy ellipse data, naturally fitting slightly smaller ellipses. In this experiments, the Kalman filter effects the results by reducing the variation amount of the observed radius, as expected.

\subsection{Object Tracking}  \label{sec:TrackingExperiments}
\begin{figure}[htb]
\begin{center}
\includegraphics*[ width=0.59 \linewidth]{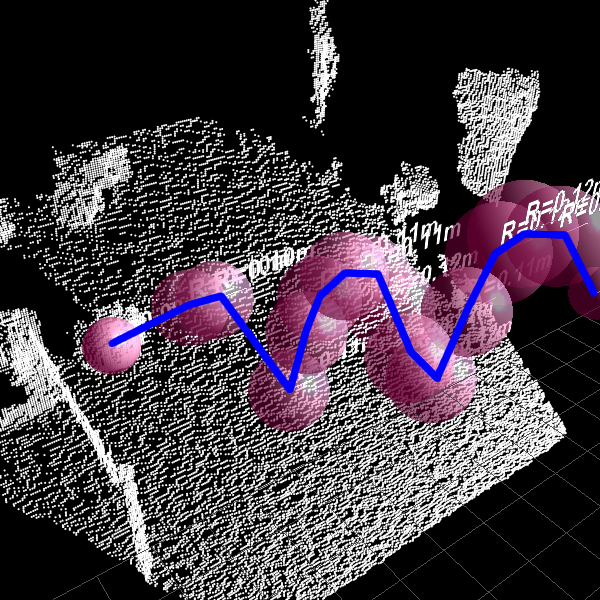}
\end{center}
\caption{\footnotesize
Results of tracking a ball. The blue line connects center points of spheres.
}
\label{fig:ball_tracing}
\end{figure}

\begin{figure}[htb]
\begin{center}
\includegraphics[ width=0.99\linewidth]{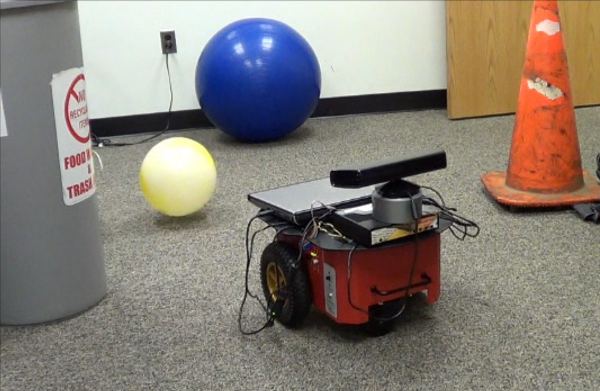}
\end{center}
\caption{\footnotesize A Pioneer DX robot tracking a ball in real-time, using 3D range data from MS Kinect.
}\label{fig:robot}
\end{figure}

This section includes three experiments. Two simple qualitative experiments, tracking a bouncing ball, and a quantitative experiment to measure object speeds.
In the first qualitative experiment, we tested if the algorithm can track a bouncing ball (radius = 10cm) see Fig. \ref{fig:ball_tracing}. Our system was able to track the ball's trajectory.
For the second qualitative experiment, we mounted a Microsoft Kinect sensor on a Pioneer mobile robot to enable navigation towards conical objects (Cylinder, Cone and Sphere). All computations were done on a notebook (2.3GHz, 2GB Ram) on the robot. The setup for the experiment is as follows: the robot has to track a ball and navigate towards it until the sphere is a meter away. In the first test the ball was held by a walking human. During the second test the ball was thrown over the robot bouncing in the desired direction. In both tests the robot detected and navigated towards the ball while moving, see Fig. \ref{fig:robot}.

\begin{figure}[htb]
\begin{center}
\includegraphics*[ width=0.99 \linewidth]{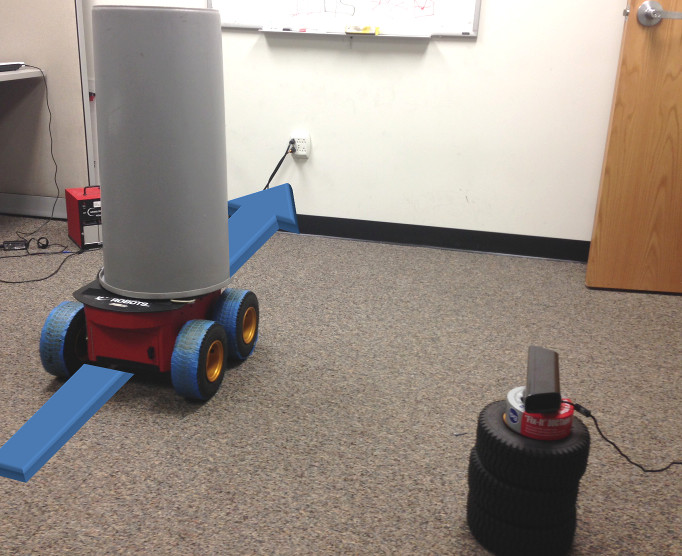}
\end{center}
\caption{\footnotesize
Experiment setup to measure speed of a moving cylinder mounted on a robot using a Kinect sensor.
}
\label{fig:robotSpeed}
\end{figure}

In the quantitative test, we placed the Kinect perpendicular to a moving cylinder on a Pioneer robot, see Fig. \ref{fig:robotSpeed} (low speed, $<=$ 1m/s ) and a moving platform dragged by a human (higher speed, $>$ 1m/s), see Table \ref{tab:speed}.
The experiment shows promising results with an increase in error at high velocity. The higher amount of error in high velocity is due to the low amount of observations (limited range of stationary Kinect). Using the Kalman filter, we could, naturally, reduce uncertainty in observations, but not necessarily improve accuracy.

%-------------------------------------------------
%-------------------------------------------------

%-------------------------------------------------
%-------------------------------------------------

%-------------------------------------------------
%-------------------------------------------------

%%%%%%%%%%%%%%%%%%%%%%%%%%%%%%%%%%%%%%%%%%%%%%%%%%%%%%%%%%%%%%%%%%%%%%%%%%%%%%%%%%%%%%%%%%%%%%%%%%%%%%%%%%%%%%%%%%%%%%%%%%%%%%%%%%%%%%%%%%%%
%-------------------------------------------------
\subsection{Comparison to RANSAC for Ellipse Fitting}\label{sec:RANSAC}
\begin{figure}[htb]
\begin{center}
\subfigure[Our Approach]{
\includegraphics[ width=0.95\linewidth]{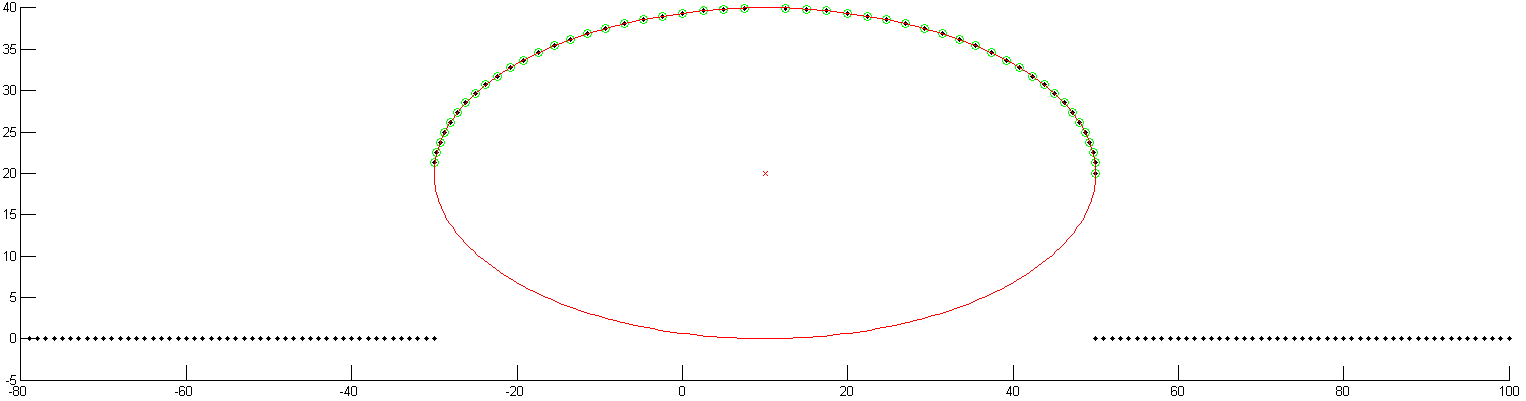}
\label{fig:RANSAC_2D_test:sub1}
}
\subfigure[RANSAC 0.2]{
\includegraphics[ width=0.95\linewidth]{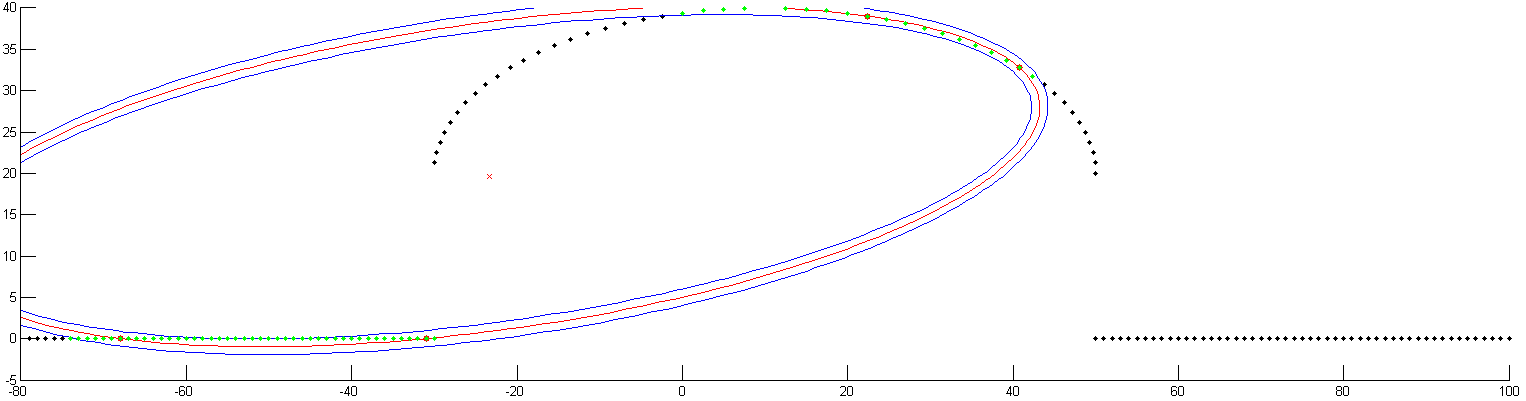}
\label{fig:RANSAC_2D_test:sub2}
}
\subfigure[RANSAC 0.4]{
\includegraphics[ width=0.95\linewidth]{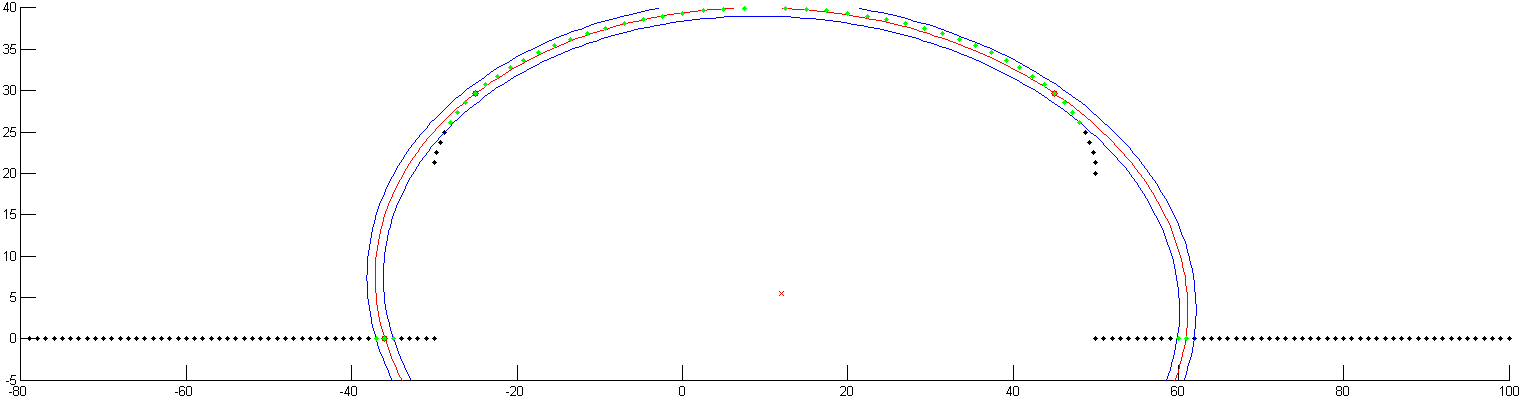}
\label{fig:RANSAC_2D_test:sub3}
}
\end{center}
\caption{\footnotesize Simulated data points (black) with fitted ellipse (red), supporting points (green). Top: Our Approach; Middle: RANSAC with constraint for radii ratio $\frac{r_{minor}}{r_{major}}\geq 0.2$; Bottom: RANSAC with stronger radii ratio constraint $\frac{r_{minor}}{r_{major}}\geq 0.2$.
}\label{fig:RANSAC_2D_test}
\end{figure}

In this section, our approach for finding ellipses (in 2D) is compared to the popular RANSAC algorithm. We perform the test on a simulated data set that represents a 2D slice of a stretched cylinder in front of a wall. We compared the results with no prior assumptions (i.e. with fixed parameters). As a stop criterion for RANSAC, we chose the number of iterations $k$ giving us a $p = 0.95$ confidence of finding the correct model. In our data set, the inlier/datapoints ratio $w$ is $w=0.3$, $n=4$ points are chosen to define an ellipse. As derived in \cite{RANSAC1981}, this results in a number of iterations is $k = \frac{log(1-p}{log(1-w^n)} < 369$. We observed two issues with RANSAC:
\begin{enumerate}
\item RANSAC is significantly slower than our approach, due to the number of attempts to find inliers(although being a $kO(n) = O(n)$ algorithm, the constant $k$ in this case is significant (in our approach, $k=1$).
\item in our data set, the outliers are not strictly randomly distributed, but they are structured (wall), which is a valid assumption for indoor and outdoor environments. Without restricting RANSAC to specific ellipses with constrained ratio of radii (minor/major axis), it finds ellipse support in the outliers (in our case, the linear structures), leading to degenerated results, see Figure \ref{fig:RANSAC_2D_test:sub1}, center. This is a RANSAC-inherent problem, since the approach does not take into account data point neighborhood constraints. To constrain RANSAC to choose points from smaller neighborhoods leads in turn to less robustness to noise. In comparison, our approach is able to find ellipses with minimal constraints on ratio of radii, and is more robust to noise. Naturally, RANSAC improves when the ratio of radii is specified with stronger constraints, see Figure \ref{fig:RANSAC_2D_test:sub1}, bottom.
\end{enumerate}
In summary, the basic RANSAC approach is less feasible for the purpose of finding ellipses in data for reasons of real time performance and parameter flexibility.

%====================================================================
\section{Conclusion and Future Work}  \label{sec:conclusion}
%====================================================================
We presented an algorithm for real-time extraction of conical objects from 3D range data using a three step approach (points to ellipses to objects). It provides fast and robust performance for extracting all 2D best fit ellipses in $O(n)$, $n$ number of points. The algorithm is well behaved towards noise and can aid higher level tasks, for example, autonomous robot navigation by providing robust and fast landmark features. To improve robustness, the algorithm can be augmented using a Kalman filter, which is feasible due to the algorithm's fast performance. Future work consist of improving the algorithm to predict models when conical objects are significantly occluded.
%====================================================================

{\small
\bibliographystyle{ieee}
\bibliography{references} % the bib file is there
}

\end{document}